\documentclass[12pt]{article} 
\usepackage{graphicx} % Required for inserting images
\usepackage{graphicx}
\usepackage[margin=1in]{geometry}
\usepackage{array} 
\usepackage{multirow}
\usepackage{booktabs}
\usepackage{amssymb}
\usepackage{amsmath}
\usepackage{url}
\usepackage{tabularx}

\title{Leaving Some Facial Features Behind}
\author{Cheng Qiu \\ Vanderbilt University}
\date{}

\begin{document}

\maketitle

\begin{abstract}
Facial expressions are crucial to human communication, offering insights into emotional states. This study examines how specific facial features influence emotion classification, using facial perturbations on the Fer2013 dataset. As expected, models trained on data with the removal of some important facial feature experienced up to an 85\% accuracy drop when compared to baseline for emotions like happy and surprise. Surprisingly, for the emotion disgust, there seem to be slight improvement in accuracy for classifier after mask have been applied. Building on top of this observation, we applied a training scheme to mask out facial features during training, motivating our proposed Perturb Scheme. This scheme, with three phases—attention-based classification, pixel clustering, and feature-focused training, demonstrates improvements in classification accuracy. The experimental results obtained suggests there are some benefits to removing individual facial features in emotion recognition tasks.
    
\end{abstract}
\section{Introduction}
Understanding emotions plays a pivotal role in how we perceive and interact with others. When interpreting facial emotions, key landmarks such as the eyes and mouth provide significant insights into a person's emotional state \cite{el_boudouri_emonext_2023}. Additionally, faces can be divided into two halves, focusing on the eyes and eyebrows on one side and the mouth on the other \cite{mukhiddinov_masked_2023}. To enhance our comprehension of facial emotions, it is essential to understand how these facial landmarks contribute to emotion prediction.

To investigate the impact of facial landmarks on emotion classification, we introduced perturbations such as a physical mask to Fer2013, a facial emotion dataset, through similar procedure to that of \cite{han_masked_2023}. In this paper, the masked facial emotion dataset will be referred to as MaskFer\footnote{Dataset can be accessed at \url{https://github.com/chengq220/MaskFerDataset}}. Models were trained on both Fer2013 and MaskFer. The performance results for accuracy are presented in Table \ref{tab:performance_class_pre}. The accuracy ratio for each emotion class generally decreased for model trained on MaskFer when compared to model trained on Fer2013. This is to be expected because by applying a mask, important features are removed from the context, resulting in less information models to determine the emotion. However, the percent change in accuracy ratio varied across different classes. For example, the accuracy for emotions like happiness dropped by around 60\% while for fear, there was only a slight drop of around 10\%. Surprisingly, for the emotion sad, there's even a slight increase in accuracy ratio which means that the model trained on MaskFer performed better than its counterpart trained on Fer2013. For this specific emotion, we hypothesized that masking the mouth allowed the model to extract better information relevant to the emotion of sad such as the eyebrow.

\begin{table}[h!]
    \centering
    \begin{tabularx}{\linewidth}{lXXXXXX}
        \toprule
        \small{Emotion} & \cite{huang_densely_2018} & \cite{chen_dual_2017} & \cite{pham_facial_2021} & \cite{he_deep_2015} & \cite{simonyan_very_2015} & \small{Ensemble} \\
        \midrule
        \small{Angry}      & \small{-39\%} & \small{-35\%} & \small{-32\%} & \small{-20\%} & \small{-23\%} & \small{-32\%} \\
        \small{Happy}      & \small{-49\%} & \small{-63\%} & \small{-42\%} & \small{-55\%} & \small{-35\%}   & \small{-42\%} \\
        \small{Sad}        & \small{-12\%}  & \small{-24\%}  & \small{-4\%}  & \small{-16\%}    & \small{-8\%}  & \small{-4\%}  \\
        \small{Neutral}    & \small{-51\%}   & \small{-60\%} & \small{-53\%} & \small{-60\%} & \small{-43\%} & \small{-53\%} \\
        \small{Surprise}   & \small{-19\%} & \small{-26\%} & \small{-35\%} & \small{-19\%} & \small{-20\%} & \small{-35\%} \\
        \small{Disgust}    & \small{-49\%} & \small{-59\%} & \small{-39\%} & \small{-49\%} & \small{-49\%} & \small{-39\%} \\
        \small{Fear}       & \small{-20\%}   & \small{-23\%} & \small{-13\%} & \small{-18\%}  & \small{-17\%} & \small{-13\%} \\
        \bottomrule
    \end{tabularx}
    \caption{Percent accuracy gain/loss for each emotion class with models trained on MaskFer compared to models trained on FER2013}
    \label{tab:performance_class_pre}
\end{table}

% \begin{table}[h!]
%     \centering
%     \begin{tabular}{lccccccc}
%         \toprule
%         \textbf{Model} & \textbf{Angry} & \textbf{Happy} & \textbf{Sad} & \textbf{Neutral} & \textbf{Surprise} & \textbf{Disgust} & \textbf{Fear} \\
%         \midrule
%         DenseNet \cite{huang_densely_2018} & -39.1\% & -48.5\% & 12.2\% & -51\% & -18.8\% & -49.4\% & -20\% \\
%         DPN \cite{chen_dual_2017} & -34.8\% & -62.8\% & 23.9\% & -59.6\% & -26.4\% & -58.72\% & -22.8\% \\
%         ResMasking \cite{pham_facial_2021}& -31.6\% & -42.2\% & -4.3\% & -52.7\% & -34.5\% & -39.1\% & -12.6\% \\
%         ResNet \cite{he_deep_2015} & -19.9\% & -54.6\% & 16\% & -59.7\% & -19.3\% & -49.2\% & -18\% \\
%         VGG16 \cite{simonyan_very_2015} & -22.8\% & -35\% & -7.9\% & -43.2\% & -20.1\% & -49.3\% & -17.2\% \\
%         Ensemble & -31.6\% & -42.2\% & -4.3\% & -52.7\% & -34.5\% & -39.1\% & -12.6\% \\
%         \bottomrule
%     \end{tabular}
%     \caption{Percent accuracy gain/loss models of model trained on MaskFer compared to model trained on Fer2013}
%     \label{tab:performance_class_pre}
% \end{table}

\begin{figure}[h]
    \centering
    \includegraphics[scale=0.8]{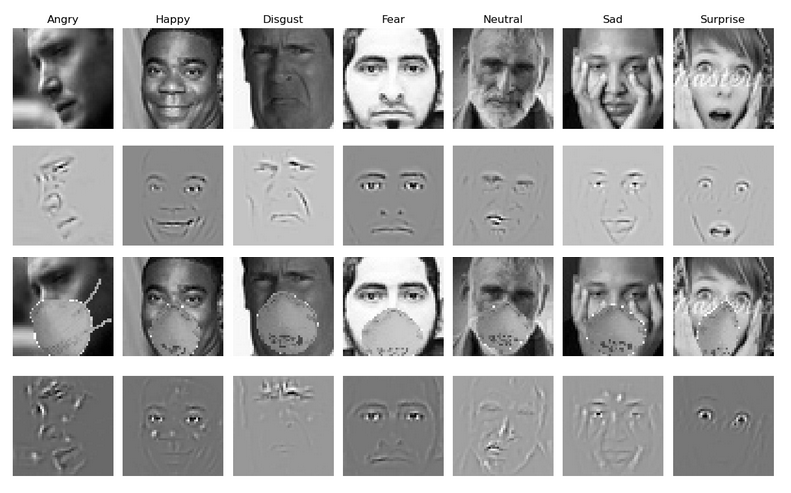}
    \caption{Saliency map for trained models on Fer2013 and MaskFer}  
    \label{fig:saliency1}
\end{figure}

By examining the saliency map shown in Figure \ref{fig:saliency1}, it becomes evident that for emotions such as disgust, anger, and sadness, the model trained on Fer2013 may not have learned the optimal salient features. In contrast, the model trained on MaskFer was able to capture the features in the eyebrows when identifying an angry face. It might be case that models trained on Fer2013 do not effectively leverage all the information present in the vicinity of major facial landmarks like the eyebrows. While classifying emotions based on overall facial landmarks may provide a global context that's beneficial to emotion classification, considering them as individual features provide better localized context that's better suited for classification of specific emotions. 

\section{Related Works}
Neural networks have become a natural choice for facial emotion recognition tasks due to their ability to learn complex representations from visual data. With the rise of deep learning, networks like ResNet \cite{he_deep_2015} have paved the way for deeper architectures by introducing skip connections, which allow information to flow across layers without degradation. This innovation enables deeper networks that can learn richer features, albeit with increased computational complexity. DenseNet \cite{huang_densely_2018} later addressed this issue by establishing direct connections between layers, effectively reducing the number of parameters while maintaining performance, which is crucial for real-time applications.

Recent advances have also seen the development of the Dual Path Network (DPN) \cite{chen_dual_2017}, which combines the advantages of both ResNet’s and DenseNet’s connectivity patterns. By incorporating a dual-path structure, DPN allows features to be reused efficiently across layers, further enhancing classification accuracy while managing computational load. The FER2013 benchmark dataset has been widely used to train and evaluate facial emotion recognition models. Many studies have built upon convolutional neural networks (CNNs) to improve baseline models. For example, the addition of attention mechanisms has enabled models to selectively focus on significant facial areas, such as the eyes and mouth, enhancing the interpretability and accuracy of emotion recognition \cite{minaee_deep-emotion_2019}. Attention mechanisms, inspired by human visual perception, help networks prioritize crucial regions that are most indicative of emotions, particularly under challenging conditions where parts of the face might be occluded or not visible.

Additionally, researchers have experimented with hyper-parameter tuning using genetic algorithms to optimize hyper-paramaters of model such as VGG architecture. This approach improved the FER2013 classification accuracy by refining aspects like kernel size and learning rate, ultimately achieving a single-network classification accuracy of 76\% \cite{saman_convolutional_2024}. Spatial transformations have also been applied to increase robustness against rotations, translations, and other variations, which are common in real-world scenarios \cite{el_boudouri_emonext_2023}.

Despite these improvements, emotion recognition models still face substantial challenges, particularly when deployed in noisy, ambiguous, or complex environments. Issues such as inconsistent annotations and the diverse real-world backgrounds can impact performance significantly. Furthermore, in applications where parts of the face are occluded, like mask-wearing scenarios, critical facial landmarks might be hidden, presenting additional challenges. Emotion classification under partial face occlusion requires models to make predictions based on incomplete information, often resulting in lower accuracy.

To address these challenges, researchers have turned to techniques like transfer learning, where models pre-trained on large, general-purpose datasets are fine-tuned on smaller, task-specific datasets. This approach has shown promising results, as pre-trained models can leverage learned representations from a broad data distribution to perform well on specific tasks, even with limited data \cite{huang_study_2023}. Moreover, the introduction of new datasets designed to include occlusions, such as MaskFer, enables models to better generalize to masked or partially visible faces. MaskFer, for instance, includes images with the lower part of the face masked, simulating scenarios in which only the upper facial features are visible. Research utilizing CNNs in these scenarios aims to identify and adapt to visible facial landmarks, acknowledging that masking impacts emotion recognition by obscuring expressive regions like the mouth and nose \cite{mukhiddinov_masked_2023}.

Innovative approaches specifically for masked faces have leveraged landmark-based and low-light image enhancements. One study applied histogram-based regional representation and boundary analysis to focus on the upper face, using AffectNet, a comprehensive dataset containing over 420,000 images across eight emotion classes. The method involved covering the lower half of the face with synthetic masks, then extracting features from the head and upper facial regions, such as the eyes and eyebrows. By combining facial landmark detection with histograms of oriented gradients (HOG), the approach achieved effective classification with CNNs, even under occlusion\cite{mukhiddinov_masked_2023}.

\section{Methodology}
\begin{figure}[h]
    \centering
    \includegraphics[scale=0.6]{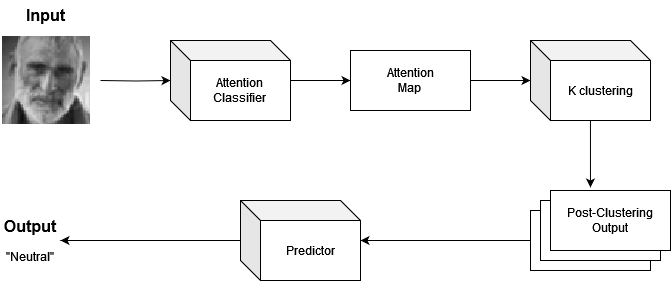}
    \caption{Framework for Perturb Scheme}
    \label{fig:model3}
\end{figure}
To fully leverage the regional characteristic in facial recognition, we propose a new scheme, the Perturb Scheme, shown in Figure \ref{fig:model3} which includes threes phases. The first phase focuses on training an attention classifier to better pin point the important pixels. The second phase involve clustering those pixels into classes base on importance. Thirdly, training a new classifier to perform the emotion classification task with the inputs from the clustering. Refer to Figure. \\

\noindent \textbf{Attention Classifier: } The first phase is training a neural network focused on understanding the spatial attention in order to better understand the influence of each pixel toward the prediction of all emotion. Since different facial landmarks may have different effects on the performance of emotion predictions. To achieve this, spatial attention modules were added between layers of the neural networks \cite{woo_cbam_2018}.\\

\noindent \textbf{Clustering: } The second phase is to isolate local patches of pixels that have the highest attention. To identify those patches, we will use K-mean clustering with $n$ clusters to represent $n$ meaningful facial landmarks. By leveraging the attention, pixels is clustered into different classes to be masked out based on the euclidean distance shown in Equation \ref{eq:euclid}. Due to the computation required during clustering, it's infeasible to perform clustering for each image, instead the cluster updates every epoch. \\

\noindent Given an $I \in {\rm I\!R}^{BXCXWXH}$ and its spatial attention map $S \in {\rm I\!R}^{BXWXH}$, the distance between any two points on the 2D grid $P_{ij}$ and $P_{kl}$, where $i,k \in W$, $j,l \in H$. The Euclidean distance can be computed as shown in Equation \ref{eq:euclid} where $\lambda$ and $\alpha$ are constants used to control the weights of the pixel distance on the grid and the intensity distance.

\begin{equation}
    \text{Dist} = \sqrt{(\lambda(i-k))^2 + (\lambda(j-l))^2 + (\alpha(I_{ij} - I_{jl}))^2} \label{eq:euclid}
\end{equation}

\noindent \textbf{Predictor Training: } The third phase is training a new network to learn from the clustered data in order to optimize the emotion detection task.

\section{Experiment}
We empirically demonstrate the effectiveness of the Perturb Scheme in enhancing the capabilities of deep learning models for emotion recognition. Our performance comparisons include prominent deep learning architectures such as DenseNet, ResNet, and others.

\subsection{Dataset} 
\begin{figure}[h]
    \centering
    \includegraphics[scale=0.3]{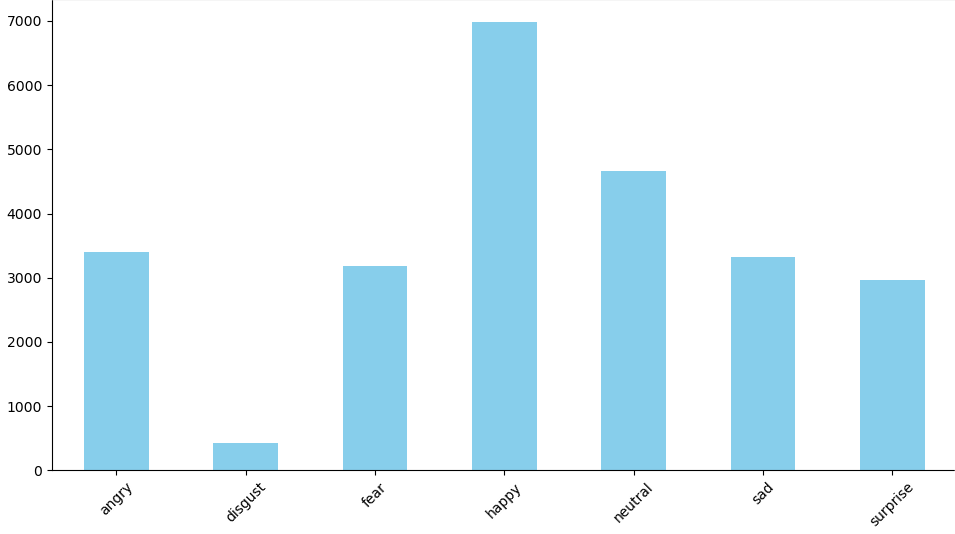}
    \caption{Distribution of classes in Fer2013 dataset}
    \label{tab:dist}
\end{figure}

FER2013 is a publicly available well-known dataset for facial emotion recognition. The dataset consists of 35887 gray scaled images of emotions The FER2013 dataset, a well-known publicly available resource for facial emotion recognition, consists of 35,887 gray scale images across seven different emotion classes. Of these, 28,708 images are used for training, while 3,589 images are designated for validation and testing, making the dataset suitable for deep learning applications. However, FER2013 contains incorrectly labeled faces, irrelevant images, and an imbalanced distribution of image data, with some classes containing up to 7,000 images and others as few as 500 (see Figure \ref{tab:dist}). Do take note that since MaskFer is directly built on top of FER2013, similar issues exist in MaskFer.
\subsection{Training}
All models were trained on an NVIDIA RTX 3090 for 200 epochs using cross-entropy loss as the primary loss function. Stochastic Gradient Descent (SGD) with Nesterov momentum (0.9) and weight decay was employed. The initial learning rate was set to 0.01 and adjusted during training using ReduceLROnPlateau (RLRP) \cite{pramerdorfer_facial_2016}. For data preparation, we adopted the image augmentation techniques recommended by Zhong et al., including random flipping, cropping, and erasing, to prevent over-fitting, especially given the scarcity of samples in some classes \cite{zhong_random_2017}. For the Perturb Scheme, the parameters $\alpha$ and $\lambda$ in Equation \ref{eq:euclid} were set to 1.2 and 1.5 respectively with the number of clusters for the k-mean clustering set to 3. \\

\section{Evaluation and Analysis}
For evaluation, all models were trained on FER2013 and tested using the testing set from the same dataset. We selected several models commonly used for emotion classification, including ResMasking \cite{pham_facial_2021} and ResNet \cite{he_deep_2015}, as well as baseline models like VGG16 \cite{simonyan_very_2015}, to assess the performance of the proposed Perturb Scheme. Our results demonstrate that the proposed method does yield positive outcomes.

\subsection{Attention-Based Clustering}

\begin{figure}[h]
    \centering
    \includegraphics[scale=0.8]{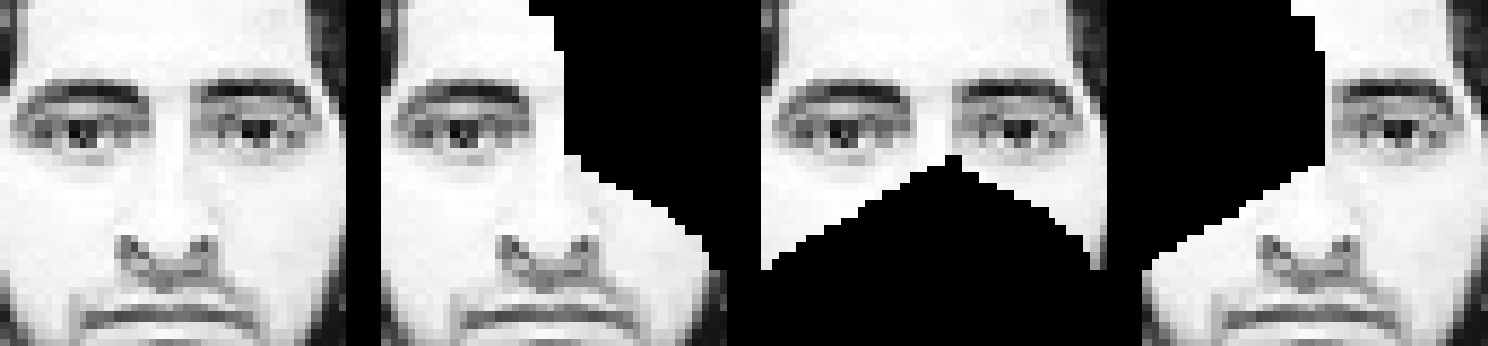}
    \caption{Facial image after applying clustering $n=3$ based on the attention}
    \label{fig:filter}
\end{figure}

\begin{figure}[h]
    \centering
    \includegraphics[scale=0.55]{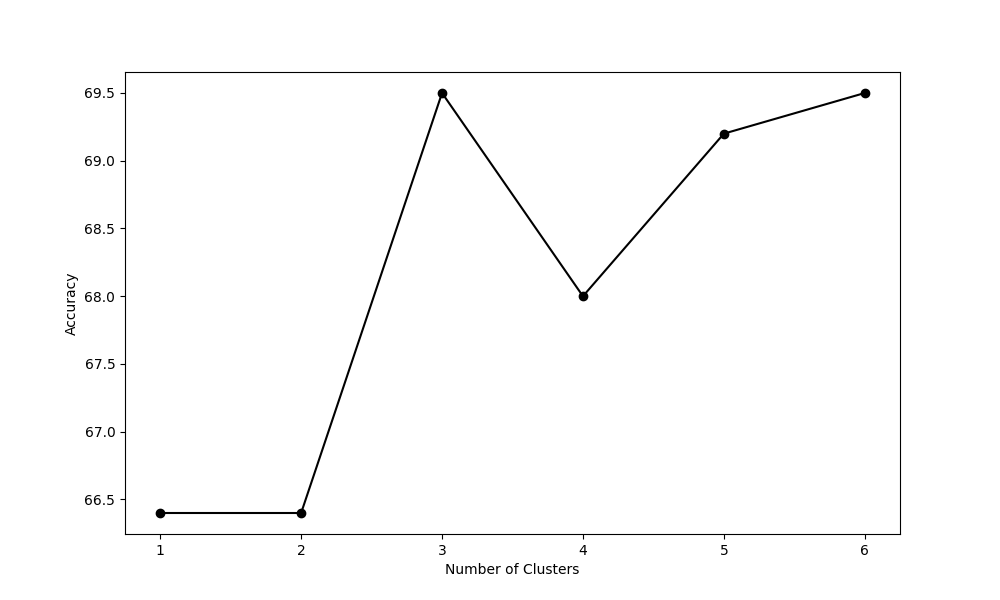}
    \caption{Performance for different number of clusters for the VGG16 architecture}
    \label{fig:clust}
\end{figure}

\noindent In spatial attention-based emotion classification, our models primarily focus on localized regions around the eyes and mouth, as highlighted by the attention classifier. This approach underscores the importance of these facial areas, with two clusters targeting the eyes and one focusing on the mouth and nose region. As shown in Figure \ref{fig:clust}, we experimented with varying the number of clusters, finding that three clusters yielded optimal performance by isolating the most relevant features for emotion recognition. It matches up with the intuition that generally on face there's 3 salient landmarks which are the two eyes and the mouth. Furthermore, adding more clusters increased computational demands without consistent improvements in accuracy, and in some cases, led to performance decline.
\subsection{Performance}
\begin{table}[htbp]
    \centering
    \begin{minipage}{.50\linewidth}
    \centering
    \textbf{2.1 Summary (Baseline)}
        \begin{tabular}{lcc}
        \toprule
        \textbf{Model} & \textbf{Accuracy} & \textbf{F1} \\
        \midrule
        \cite{huang_densely_2018} & 72.6\% & 0.70 \\
        \cite{chen_dual_2017} & 71.2\% & 0.68 \\
        \cite{pham_facial_2021} & 66\% & 0.63 \\
        \cite{he_deep_2015} & 69.2\% & 0.685 \\
        \cite{simonyan_very_2015}& 66.5\% & 0.653\\
        Ensemble & 73.4\% & 0.71 \\
        \bottomrule
    \end{tabular}
    \end{minipage}%
    \hfill
    \begin{minipage}{.50\linewidth}
    \centering
    \textbf{2.2 Summary (Perturb)}  
        \begin{tabular}{lcc}
        \toprule
        \textbf{Model} & \textbf{Accuracy} & \textbf{F1} \\
        \midrule
        \cite{huang_densely_2018} & -- & -- \\
        \cite{chen_dual_2017}& 71.9\% & 0.69 \\
        \cite{pham_facial_2021} & -- & -- \\
        \cite{he_deep_2015} & 72.3\% & 0.70 \\
        \cite{simonyan_very_2015} & 69.5\% & 0.66 \\
        Ensemble & --\ & -- \\
        \bottomrule
    \end{tabular}
    \end{minipage}%
    \caption{Table 2.1 and Table 2.2 illustrate the performance comparison between classifier trained without and with Perturb Scheme}
    \label{tab:exp1}
\end{table}

\begin{figure}[h]
    \centering
    \includegraphics[scale=0.65]{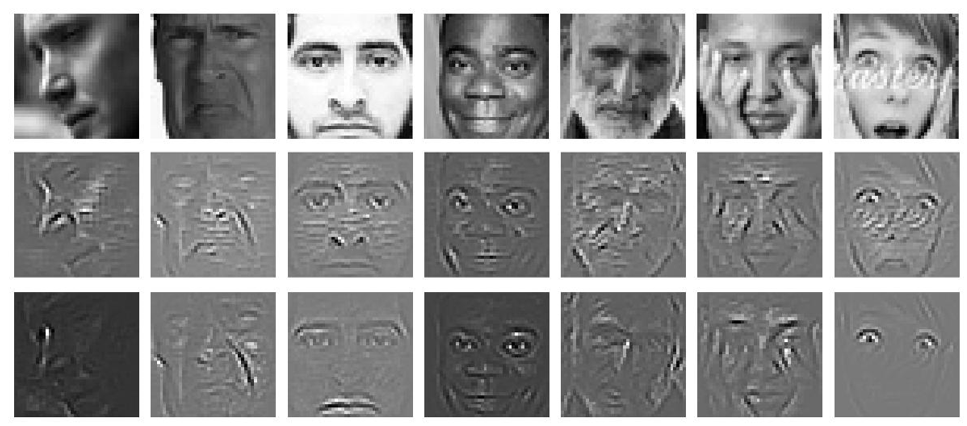}
    \caption{Comparison of the saliency map between VGG16 and VGG16 with Perturb Scheme (VGG16 in row 2 and VGG16 with Perturb Scheme in row 3)}
    \label{fig:rsalient}
\end{figure}

 Limited by computation resources, only a selected few of the proposed architecture were trained using Perturb Scheme. The models trained using Perturb scheme generally performed better in accuracy cross all emotion class as shown in Table \ref{tab:performance_class}. It further suggests that removing facial landmarks such as mouth and eyes in training did contribute to the increase in accuracy. Comparing the saliency map for the baseline model and the model trained using proposed method shown in Figure \ref{fig:rsalient}, it can be observed for emotions such as fear, the new models are able to put more emphasis on the important features such as the mouth and less emphasis on less important features like the nose. 

\begin{table}[h!]
    \centering
    \begin{tabularx}{\linewidth}{lXXXXXX}
        \toprule
        \small\textbf{Emotion} & \cite{huang_densely_2018} & \cite{chen_dual_2017} & \cite{pham_facial_2021} & \cite{he_deep_2015} & \cite{simonyan_very_2015} &\small{Ensemble} \\
        \midrule
        \small{Angry}     & --    & \small{0.8\%}  & --    & \small{3.5\%}   & \small{12\%}   & --    \\
        \small{Happy}     & --    & \small{2.0\%}     & --    & \small{1.0\%}    & \small{-1\%}   & --    \\
        \small{Sad}       & --    & \small{1.5\%}   & --    & \small{1.5\%}   & \small{-17\% } & --    \\
        \small{Neutral}   & --    & \small{1.4\%}   & --    & \small{-0.6\%}  & \small{2.2\% }   & --    \\
        \small{Surprise}  & --    & \small{-0.5\%}  & --    & \small{2.1\%}   & \small{1.6\%}    & --    \\
        \small{Disgust}   & --    & \small{-3.6\%}  & --    & \small{5.6\% }  & \small{5.0\%}      & --    \\
        \small{Fear}     & --    & \small{3.0\%}     & --    & \small{8.4\% }  & \small{18\% }    & --    \\
        \bottomrule
    \end{tabularx}
    \caption{Changes in performance for each emotion class with the perturb scheme on FER2013 dataset}
    \label{tab:performance_class}
\end{table}

% \begin{table}[h!]
%     \centering
%     \begin{tabular}{lccccccc}
%         \toprule
%         \textbf{Model} & \textbf{Angry} & \textbf{Happy} & \textbf{Sad} & \textbf{Neutral} & \textbf{Surprise} & \textbf{Disgust} & \textbf{Fear} \\
%         \midrule
%         DenseNet \cite{huang_densely_2018} & -- & -- & -- & -- & -- & -- & -- \\
%         DPN \cite{chen_dual_2017} & 0.8\% & 2\% & 1.5\% & 1.4\% & -0.5\% & -3.6\% & 3\%\\
%         ResMasking \cite{pham_facial_2021}& -- & -- & -- & -- & -- & -- & -- \\
%         ResNet \cite{he_deep_2015} & 3.5\% & 1\% & 1.5\% & -0.6\% & 2.1\% & 5.6\% & 8.4\% \\
%         VGG16 \cite{simonyan_very_2015} & 12.2\% & -1.2\% & -17.3\% & 2.2\% & 1.6\% & 5\% & 18\% \\
%         Ensemble & -- & -- & -- & -- & -- & -- & -- \\
%         \bottomrule
%     \end{tabular}
%     \caption{Changes in class-wise performance of models trained using perturb scheme over the baseline}
%     \label{tab:performance_class}
% \end{table}

\section{Conclusion}
Our findings demonstrate that incorporating regional feature extraction through the Perturb Scheme significantly enhances emotion recognition model performance, especially under occlusion conditions. By applying attention-based clustering and subsequently training on clustered data, our approach enables models to prioritize the most informative facial landmarks, leading to marked improvements in predictive accuracy across most emotion classes. Notably, the slight decline in accuracy for emotions like disgust suggests that additional refinements may be beneficial for capturing subtle expressions tied to specific regions of the face.

The Perturb Scheme's ability to enhance model robustness against occlusions underscores the value of regional feature emphasis in facial emotion recognition tasks. This approach has potential applications in real-world settings where full facial visibility cannot be guaranteed, such as masked environments or low-light conditions. These findings pave the way for future work exploring more nuanced feature emphasis strategies and adaptive techniques to optimize model accuracy across diverse expressions and contexts, contributing to the advancement of facial emotion recognition technologies.

\clearpage
\bibliographystyle{plain}
\bibliography{FacialRecognition}

\end{document}